# Current Studies and Applications of Krill Herd and Gravitational Search Algorithms in Healthcare

**Rebwar Khalid Hamad[1*] and Tarik A. Rashid[2]**

[1]Department of Information Systems Engineering, Erbil Technical Engineering College, Erbil Polytechnic University, Erbil, Iraq

[2]Computer Science and Engineering Department, University of Kurdistan Hewler, Erbil, Iraq

*Corresponding author: Rebwar Khalid Hamad

**Abstract**

Nature-Inspired Computing or NIC for short is a relatively young field that tries to discover fresh methods of computing by researching how natural phenomena function to find solutions to complicated issues in many contexts. As a consequence of this, ground-breaking research has been conducted in a variety of domains, including synthetic immune functions, neural networks, intelligence of swarm, as well as computing of evolutionary. In the domains of biology, physics, engineering, economics, and management, NIC techniques are used. In real-world classification, optimization, forecasting, and clustering, as well as engineering and science issues, meta-heuristics algorithms are successful, efficient, and resilient. There are two active NIC patterns: the gravitational search algorithm and the Krill herd algorithm. The study on using the Krill Herd Algorithm (KH) and the Gravitational Search Algorithm (GSA) in medicine and healthcare is given a worldwide and historical review in this publication. Comprehensive surveys have been conducted on some other nature-inspired algorithms, including KH and GSA. The various versions of the KH and GSA algorithms and their applications in healthcare are thoroughly reviewed in the present article. Nonetheless, no survey research on KH and GSA in the healthcare field has been undertaken. As a result, this work conducts a thorough review of KH and GSA to assist researchers in using them in diverse domains or hybridizing them with other popular algorithms. It also provides an in-depth examination of the KH and GSA in terms of application, modification, and hybridization. It is important to note that the goal of the study is to offer a viewpoint on GSA with KH, particularly for academics interested in investigating the capabilities and performance of the algorithm in the healthcare and medical domains.

# 1. Introduction

Artificial intelligence (AI) is intelligence displayed by machines rather than by people or animals. Examples of AI applications include voice recognition, intelligent agents, computer vision, and natural language processing. Across a broad array of industries, from economics to public policy to national security, AI and analytics are becoming increasingly popular as cutting-edge technologies (Liu, 2020). The healthcare sector and medical practices have seen substantial shifts as a result of AI/analytics innovation and learning algorithms. AI can affect public health, lower costs, and enhance patient outcomes in the healthcare sector. Because the amount of data being created today much exceeds the ability of human cognition to handle it effectively, AI is expected to play a crucial as well as supplementary function in assisting the provision of tailored healthcare. In regards to picture and signal recognition, for example, current AI advancements have exhibited high levels of accuracy and are considered among the most mature tools in this field (Matheny et al., 2020).

During current history, a large number of Nature-Inspired Algorithms (NIAs) have evolved. The NIA clan is rapidly growing (Kumar et al., 2022). Researchers have created a variety of nature-inspired algorithms in recent years to tackle different problems, including those in healthcare fields. Modern metaheuristic algorithms have been created and used to address these challenging issues since they were inspired by nature(G. G. Wang et al., 2019). They are all modeled on the natural behavior of bee swarms, ants, and bird flocks. The attraction of these algorithms derives from their ability to reliably and efficiently tackle NP-hard problems (González-Álvarez et al., 2013). These algorithms may be divided into two groups:

swarm-based algorithms and evolutionary algorithms (EA). Ant colony optimization, cuckoo search, firefly, and krill herd algorithms are examples of the second category of swarm-based algorithms, whereas genetic algorithms, genetic programming, and evolution strategy are examples of the first group of evolutionary algorithms.

As a result, scholars have developed many metaheuristic algorithms since 1948, when Alan Turing cracked the code of the Enigma encryption machine. The genetic algorithm (Holland, 1975), which replicates natural evolution, was devised after Turing's heuristic technique. Many methods have been created since the GA was suggested, for example, Tabu Search (Glover, 1989), Simulated Annealing (Bertsimas & Tsitsiklis, 1993), Bacterial Foraging Algorithm (BFA) (Passino, 2002), Fitness Dependent Optimizer (Abdullah & Ahmed, 2019), and Fire Hawk optimizer (FHO) (Azizi et al., 2023).

Most of us may have wondered if metaheuristic algorithms could play a role in treating and improving medical services. Can we rely on these algorithms to identify and distinguish between diseases? Is it possible to predict disease through the use of this algorithm?

The motivation of this study is to summarize knowledge of algorithms and the problems they face when using these theories, including KH and GSA. Moreover, the public has benefited from the ability to employ these algorithms in the field of healthcare. Furthermore, when contrasted with other methods and a large number of studies, they are the most powerful, commonly used, and strongest algorithms. The KH and GSA algorithms are nature-based. Another motivation is that we want to work on these two algorithms and hybridize and modify them. In addition, combine the positive aspects and limitations of those algorithms in improving versions. The main contributions of this study can be summarized as:

- A recent metaheuristics survey is presented in the paper. About two metaheuristics are included in the data set for this investigation.

- In this research, our contributions to this review are the ones that these algorithms have not been reviewed or prepared for in the field of healthcare.

- The various versions of the KH and GSA algorithms and their applications in healthcare are thoroughly reviewed in the present article.

- Above all, both algorithms have not been reviewed together. They have not been discussed in great terms.

- Our initial goal is to incorporate papers relevant to KH and GSA in healthcare during the last eight years and highlight the most current developments in the two algorithms that have been produced.

- On the other hand, the major goal of this study is to offer an outline summarizing the review of KH and GSA, such as the history of algorithms, the basic structure, mathematical works, modification, and hybridization algorithms.

The following is a breakdown of the paper's structure: First, the introduction to KH and GSA is given in Section 1, followed by a historical background in Section 2, and then a breakdown of the structure of the algorithm in Section 3. Section 4 illustrates applications in the related healthcare area, and Section 5 demonstrates modifications of the algorithms. In Section 6, the hybridization of algorithms is discussed, and then the conclusion and a few recommendations are given.

## 2. Research Methodology

The evaluation of KHA and GSA followed the Preferred Reporting Items for Systematic Reviews and Meta-Analyses (PRISMA) criteria (Moher et al., 2009). To find research publications connected to KH and GSA, a thorough search has been conducted on Scopus and Google Scholar. This document also includes significant research articles that were discovered via manual search. When searching, terms like "Krill Herd Algorithm", "KH Algorithm", "Application of KH", "Operators of KH", "Representation of KH", "Variants of KH" and "Gravitational search Algorithm", "GS Algorithm", "Application of GSA", "Operators of GSA", and "Representation of GSA". The criteria listed in Table 1 are used to decide which research articles to accept and which to reject.

**Table 1**
**The standards used to choose the selected research articles**

| # | Parameters | Criteria of selection | Criteria of elimination |
|---|---|---|---|
| 1 | Period | published studies between 2015 and 2023 | earlier than 2015, published research articles |
| 2 | Examination | Numerous operators, hybridization, and modification in KH and GSA are being studied. | Operators of different metaheuristics are used in research. |
| 3 | Applications | research that is related to healthcare | research that isn't related to healthcare |
| 4 | Study | The research incorporates both results from experiments and mathematical underpinnings. | The research includes publications written in languages other than English, case studies, and patents. |

On Google Scholar and Scopus, 2,117,733 research publications were looked up. For healthcare applications, research on the Gravitational Search algorithm and the Krill Herd algorithm was also included. All duplicate publications and those published before 2015 were eliminated during the screening of research papers. Based on 2015 entries and duplicate submissions, 2311 research articles were chosen. Following that, 1536 research articles were discarded based on their titles. 203 research articles were discarded following the abstract reading. After the third round of screening, 82 research articles remained. Following a thorough review of the articles and the discovery of facts, 50 further research papers were deleted. The final 33 research articles are chosen for assessment after the fourth round of screening.

We have taken the approach of selecting only those papers that have used the above two hypotheses or theories and have played a role in solving and improving medical problems. The purpose of this restriction was only to use recent research in the field as the primary and reliable source.

## 3. Historical Overview

This section provides the reader with a general historical overview of KH and GSA algorithms.

**2.1 History of KH**

In 2012, Gandomi and Alavi presented the Krill Herd algorithm for solving global optimization functions (Gandomi & Alavi, 2012). It is a swarm intelligence search method that draws inspiration from the krill's tendency to herd one another. The shortest distance between each krill and its nearest source of food and the herd's greatest density is used in the KH algorithm to calculate the goal function for krill migration. According to three operational processes: (A) movement caused by other individuals, (B) foraging motion, (C) and random physical diffusion—each person in the KH algorithm alters its location. Because it includes both exploration and exploitation tactics based on foraging movement and motion caused by other individuals, respectively, the KH algorithm is referred to as a strong search approach. It is regarded as one of the rapidly expanding algorithms inspired by nature that may be used to tackle real-world optimization issues.

**2.2 Research Trends of KH**

As of June 12, 2023, Scopus and Google[1] Scholar indicate that the original publication has been referenced 1,474 and 1944 times, respectively. Thus, since the KH algorithm was introduced in 2012, up until June 12, 2023, 1944 related papers have been published in conferences, journals, or dissertations. 72 papers were published in 2012-2013, 93 papers in 2014, 103 papers in 2015, 168 papers in 2016, 152 papers in 2017, 214 papers in 2018, 226 papers in 2019, 261 and 282 papers in 2020-2021, 282 papers in 2022 and 91 papers in 2023. There have been the highest number of citations in the last two years. While many papers are still in production, it is impossible to obtain all of them. For a greater understanding of the progress of the Krill Herd Algorithm from the start of this theory's discovery in 2012 to 2023, Figure 1 shows a historical chronology of the algorithm up to that point. During the 8 years and 5 months of working on the paper from 2015 to 2023, it is shown in the figure that it has been cited about 1779 times in different study fields. Moreover, KH has increased its use in the last three years, and it has only been used, modified, or hybridized with other active algorithms, and the algorithm benefited from them in the fields that had real problems. This algorithm plays a common role in solving most real-world problems in areas such as: computer science, engineering, mathematics, energy, decision sciences, etc.

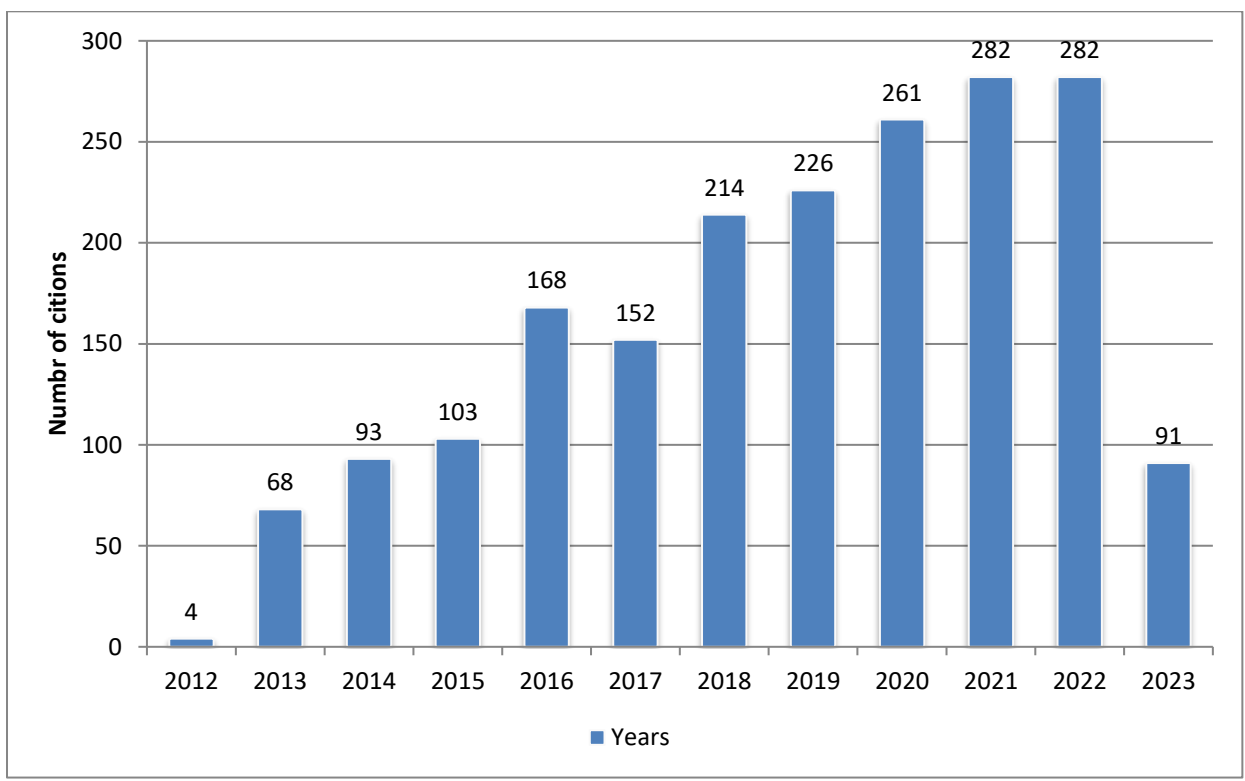

**Fig 1: Number of KH publications per year from 2012 to 2022**

**2.3 History of GSA**

A revolutionary strategy based on gravity, known as the Gravitational Search Algorithm, has been developed to address cutting-edge optimization problems. The Newtonian gravity equation defines a force that is inversely proportional to the square of the distance between two particles and directly proportional to the product of the masses of two particles (Rashedi et al., 2009). The four main conditions for each mass "agent" in GSA are position, inertia mass, active and passive gravitational mass, and mass. A fitness function is used to locate the mass and calculate its gravitational and inertial masses. The algorithm can run by changing the gravitational and inertial masses so that each mass points to a possible solution.

---

[1]https://plu.mx/plum/a/?doi=10.1016/j.cnsns.2012.05.010&theme=plum-sciencedirect-theme&hideUsage=true

### 2.4 Research Trends of GSA

It should be noted that as of June 12, 2023, the actual paper has been referenced 6936 times in Google Scholar and 5160 times in Scopus[2]. GSA was published in 2009 and from 2009 to 2023, it has been mentioned many times in related papers that have been published in conferences, journals, or dissertations. 36 papers were published in 2009-2010, 88 papers in 2011, 162 papers in 2012, 271 papers in 2013, 362 papers in 2014, 466 papers in 2015, 510 papers in 2016, 530 and 631 papers in 2017-2018, 674 papers in 2019, 802 and 926 papers in 2020-2021, 1040 papers in 2022 and 438 papers in 2023. This work focused more on the 8 years and 5 months that it has been used as a source from 2015 to 2023, which is about 6017 times. Figure 2 depicts the number of original publications published since the algorithm's initial implementation in 2009. In the three years since its origination, the number of citations has been limited. After that, related studies have been published with a dramatic increase. The most citations were in the last four years, from 2019 to 2022. Especially in 2022, there was the highest number of citations. Moreover, GSA has increased its use in the last three years, and it has only been used, modified, or hybridized with other active algorithms when the algorithm benefits from them in the fields that had real problems. The majority of problems facing fields like computer science, engineering, mathematics, energy, materials science, etc. are solved using that method.

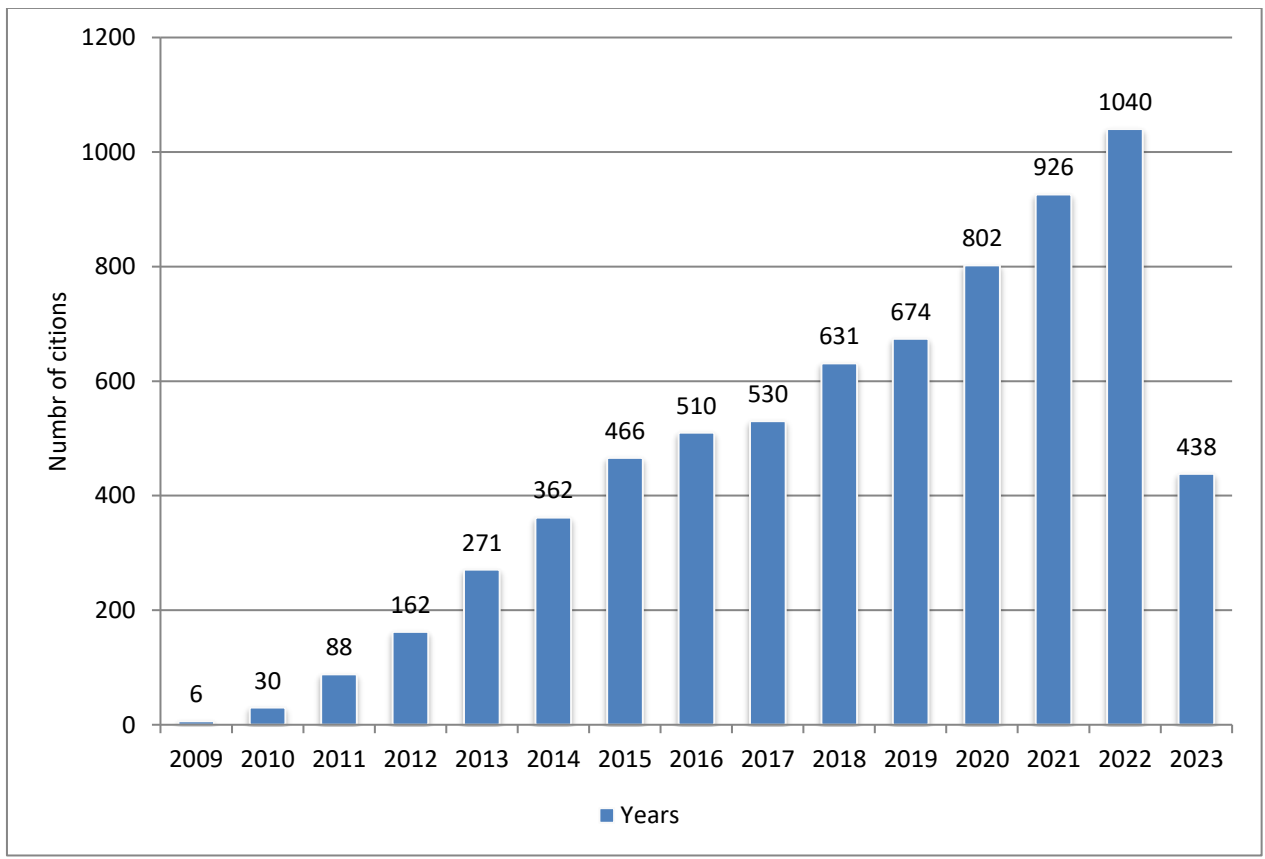

**Fig 2: Number of GSA publications per year from 2009 to 2021**

---

[2] https://plu.mx/plum/a/?doi=10.1016/j.ins.2009.03.004&theme=plum-sciencedirect-theme&hideUsage=true

**Table 2** and **Table 3** show the reference types, variants, and improvements of the algorithm, comparative research of other algorithms, the application of algorithms to real-world problems, etc.

**Table 2**
**Variants, applications, and improvements of KH**

| Ref. | Variants | Improvement | Compared to another algorithm | Applications |
|---|---|---|---|---|
| (L. M. Abualigah et al., 2018) | Combine the KH with the harmony search (HS) algorithm (Harmony-KHA) | Hybrids | Compared with KH, HS, and PSO algorithm | data clustering technique |
| (Rafi & Bharathi, 2019) | Hybrid Adaboost KNN | Hybrids | Not compared | Breast Cancer Classification |
| (Preethi, 2018) | Krill Herd-Extreme Learning Machine (ELM) Network | Classify two-dimensional MRI brain tumor pictures. | SVM, ANFIS method, ELM classifier, and neuro-fuzzy classifier | Classification and Extraction |
| (L. M. Abualigah et al., 2017) | Hybrid of KH algorithm with harmony search (HS) algorithm, namely, H-KHA | Hybrids | Orginal KH, GA, HS, PSO, ABC, and CS | data clustering |

**Table 3**
**Variants, applications, and improvements of GSA**

| Ref. | Variants | Improvement | Compared to another algorithm | Applications |
|---|---|---|---|---|
| (Dash et al., 2015) | Fuzzy MLP-GSAPSO | Hybrids | Fuzzy MLP-GS and Fuzzy MLP-PSO | Classification of medical data |
| (Shirazi & Rashedi, 2016) | Hybrid support vector machine with Mixed Gravitational search algorithms( MGSA-SVM) | Hybrids | Compared with the SVM classifier | Classification |

| (Ezzat et al., 2021) | Hybrid CNN and GSA | Hybrids | SSD-DenseNet121 | Classification chest X-ray images |
|---|---|---|---|---|
| (Sampathkumar et al., 2022) | Combine GSOA and DBN-CNNs | Hybrids | GA, ACO, HA and WOA | Classification and Prediction |
| (Chao et al., 2018) | Hybrid GSA and EBPA | Hybrids | By contrasting the EBPA and GSA's performances | Classification CT scan images |
| (Rajesh Sharma & Marikkannu, 2015) | Hybrid RGSA and SVM | Hybrids | in comparison to a normal BPN and KNN classifier | Classification and Extraction |
| (Shukla et al., 2020) | Combine TLBO and GSA | Hybrids | TLBO and GSA | Classification and Extraction |
| (Choudhry et al., 2022) | Hybrid Enhaced GSA and Deep learning | Hybrids | ELM, MLP, LSTM, SGD and ACO | Identification and classification |

# 4. The general structure of Krill Herd and Gravitational Search Algorithm

This portion has placed a strong emphasis on the algorithms' structures.

### 3.1. Krill herd algorithm

The KH approach simulates krill herding behavior. The shortest distances between each individual krill and the herd's densest point define the target function for krill migration. The foraging habits of an individual krill, the availability of other krill, and stochastic diffusion all have an impact on its time-dependent position.

Gandomi and Alavi proposed the krill herd strategy for overall optimization. The Krill swarming behavior has inspired this swarm intelligence heuristic algorithm. It is the shortest distance from the herd's peak density to each krill that serves as the goal of the KH algorithm for krill migration. Individuals in the KH algorithm alter their location based on three operational process motions: foraging, movement, and random physical diffusion created by other individuals. Since the KH algorithm is thought to be a good search tool because it uses both exploitation and exploration strategies that focus on how the other participants move when they are foraging. Real-world optimization problems greatly benefit from this widely used, nature-inspired computational method. This is owing to its distinct benefits in terms of simplicity, flexibility, computing efficiency, and stochastic nature, which eliminate the need for derivative information. Algorithms are classified into four types: (1) Evolutionary algorithms; (2) swarm-based algorithms; (3) physics-based algorithms; and (4) human-based algorithms (Bolaji et al., 2016). Researchers utilize the KH algorithm for a broad range of purposes and applications in many different domains. Figure 3 illustrates the KH concept.

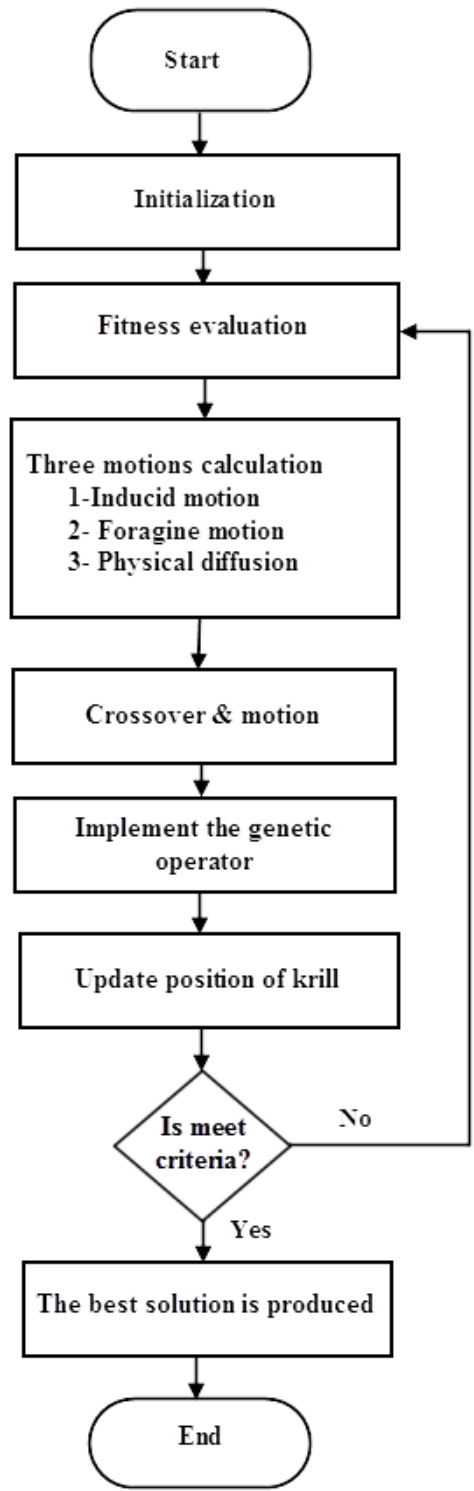

**Fig 3: The krill herd algorithm's flowchart**

Algorithm 1 displays the pseudo-code for the Krill Herd Algorithm.

```
Algorithm 1 Krill Herd Algorithm
1:   Initialization of krill parameters: V_f, RDmax, Θ^max, C_R,
     M_R, and n_p.
2:   for j = 1 to np do
3:       for i = 1 to d do
4:           xij = LBi + (UBi − LBi) × U(1, d) {Initialization
             of krill population}
5:       end for
6:       Compute f_j {Evaluate each krill}
7:   end for
8:   Sort the krill and find x^best, where best Є (1, 2,…., n_p)
9:   while t < Max_iterations do
10:      for j = 1 to np do
11:          Perform the three motion calculation using Eq.
             (1), (8), and (10)
12:          xj (t + δt) = x_j (t) + δt dx_j/dt {Update each
             krill}
13:          Fine-tune x_j+1 by using krill operators:
             Crossover and mutation
14:          Evaluate each krill by x_j+1
15:      end for
16:      Replace the worst krill with the best krill.
17:      Sort the krill and find x^best, where best Є (1, 2, . . ., n_p)
18:      t=t+1
19:  end while
20:  Return x^best
```

### 3.1.1 Category of motion calculation

In these sections, three different motion calculations are explained.

#### 3.1.1.1 Process of Motion Induced

The following equation may be used to determine the motion caused by other krill for a single krill (Gandomi & Alavi, 2012):

$$N_i^{new} = N^{max} \propto_i + \omega_n N_i^{old} \tag{1}$$

$$\propto_i = \propto_i^{local} + \propto_i^{target} \tag{2}$$

where $N^{max}$ is demonstrated by the induced maximum speed, the motion created in [0, 1] has an inertia weight of $\omega_n$, the last motion caused is $N_i^{old}$, the local effect supplied by neighbors is $\propto_i^{local}$, the best krill individual provides $\propto_i^{target}$, when is the desired direction effect.

#### 3.1.1.2 Foraging motion

Two fundamental and very effective characteristics define the foraging motion. The first is where the food is located, and the second is the dining experience (Gandomi & Alavi, 2012). The *ith* krill individual may describe this movement thusly:

$$Fi = V_f \beta_i + \omega_f F_i^{old} \tag{3}$$

$$\beta_i = \beta_i^{food} + \beta_i^{best} \tag{4}$$

where $V_f$ is the speed of foraging, $\omega_f$ represents the inertia weight of the foraging motion, which ranges from 0 to 1. And the final foraging movement is $F_i^{old}$. $\beta_i^{food}$ is the food's allure, as well as the result of the *ith* krill's finest fitness yet.

### 3.1.1.3 Physical diffusion

Individual krill are assumed to be dispersed at random. (Gandomi & Alavi, 2012) say that the motion can be defined by both a maximum speed of diffusion and a random direction vector. This may also be phrased as follows:

$$D_i = D^{max} \delta \tag{5}$$

where in eq. (3) $D^{max}$ showed the maximum diffusion speed and is a randomized direction vectors with matrices holding randomized values within the range (-1, 1).

### 3.1.2 Variants of KH

Table 4 shows many versions of the Krill Herd algorithm. The details of each of these tactics are provided below.

**Table 4**
**Variants of KH**

| Variants | Explanation | Years | Ref. |
|---|---|---|---|
| **Discrete KH** | This is the most recent version of KH that can be used for the problem of discrete optimization. | 2014 | (Sur & Shukla, 2014) |
| | This approach is also used to tackle search and optimization issues based on graph networks. | 2015 | (G. G. Wang et al., 2016) |
| **Binary KH** | In many datasets, this technique is used to address the feature selection issue. | 2014 | (Rodrigues et al., 2014) |
| **Fuzzy KH** | They proposed a fuzzy KH and used a probabilistic model like a variable tuner to adjust the amount of global and local searches. | 2014 | (Fattahi et al., 2014) |
| | A fuzzy KH (FKH) may dynamically change the volume of investigation and exploitation based on how well the | 2016 | (Fattahi et al., 2016) |
| | problem is being solved at each stage. | | |
| **Multi-objective KH** | A unique multiobjective KH (MKH) approach was introduced to address problems with multiobjective optimization. | 2016 | (Ayala et al., 2016) |

### 3.1.2.1 Discrete KH

(Sur & Shukla, 2014) proposed the first upgrade version of KH that can be used to solve the discrete optimization issue. In a different research, this strategy was developed and is also utilized to handle search and optimization issues based on graph networks (G. G. Wang et al., 2016)

### 3.1.2.2 Binary KH

The binary KH method was suggested by (Rodrigues et al., 2014). On many datasets, the approach is used to address the feature extraction challenge.

### 3.1.2.3 Fuzzy KH

A fuzzy KH and using a probabilistic model like a parametric tuner to regulate the number of global and local searches were proposed by (Fattahi et al., 2014). Additionally, in a different paper (Fattahi et al., 2016), the authors developed a fuzzy KH (FKH) that could dynamically alter the degree of exploration and exploitation following how well the issue was being solved at each stage.

### 3.1.2.4 Multi-objective KH

To solve issues involving optimization techniques, (Fattahi et al., 2016) introduced a special multiobjective KH (MKH) method. The beta distributions are employed to calculate the inertia weight change in the modified MKH technique.

### 3.2. Gravitational search algorithm

A population search algorithm was used to define the Gravitational Search Algorithm in 2009, according to (Rashedi et al., 2009). The Gravitational Search Algorithm, depending on Newton's law, is one of the most recent meta-heuristic algorithms. Each of the universe's particles continues to attract any other particle with such a force that really is directly related to the product of the active mass of the exerting particle by the passive mass of the experiencing particle, and proportional to the inverse of the distance between them. This is identified as Newton's Theory of Gravity, which forms the basis of the (GSA). The acceleration that results when a pressure is acted to an object relies on both the applied force and the mass of the item's inertial component. GSA has a similar structure to Particle Swarm Optimization (PSO). The moderate interactions between mass and gravity are used to build the GSA. Agents are the GSA population's solutions, and gravity is the medium via which they communicate with one another. The population's weight and number of agents are used to evaluate performance. Objects of greater mass attract other objects of greater mass because of gravity. This stage depicts the object's global motion (exploration step), whereas the agent with a hefty mass represents the algorithm's exploitation step. The solution with the highest mass is the best option. GSA is a population-based technique that is considered easier to understand (Sabri et al., 2013). The program makes an effort to enhance a population-based algorithm that uses gravity principles in its exploitation and exploration performance. Figure 4 displays the GSA's main tasks.

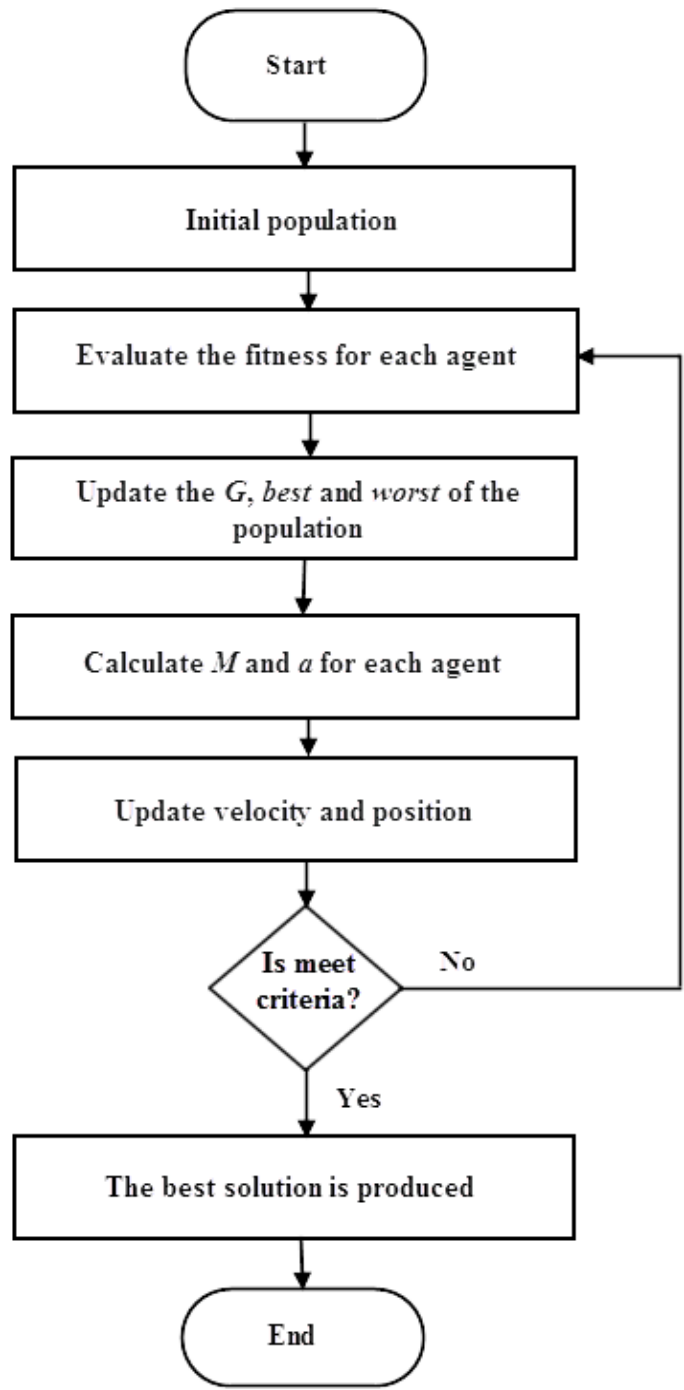

**Fig 4: A flowchart of the principle of GSA**

The pseudo-code of the Gravitational Search Algorithm is shown in Algorithm 2.

| | Algorithm 2 Gravitational Search Algorithm |
|---|---|
| 1: | Step 1: Define **initial parameters**. |
| 2: | Step 2: Create the **initial population randomly**. |
| 3: | Step 3: **While** the stop criteria are not reached |
| 4: | 3.1: Calculate the fitness function for all objects. |
| 5: | 3.2: Calculate **G, Worst, Best** |
| 6: | 3.3: Calculate the **mass** value for all objects. |
| 7: | 3.4: Calculate the **acceleration** and **velocity** of objects. |
| 8: | 3.5: Update the **position** of each object. |
| 9: | 3.6: Perform **Crossover operator**. |
| 10: | 3.7: Perform **Mutation operator**. |
| 11: | 3.8: Perform a **local search for the best solution** for this iteration |
| 12: | based on Fig. 4. |
| 13: | **End While.** |

Consider a system that consists of N agents (masses). We establish the ith agent's position as follows:

$$X_i = (x_i^1, \ldots, x_i^d, \ldots, x_i^n) \; for \; i = 1,2,\ldots,n, \tag{6}$$

Where $x_i^d$ displays the *ith* agent's location in the *dth* dimension.

We define the force exerted on mass *'i'* from mass *'j'* at a certain time *'t'* as follows:

$$F_{ij}^d(t) = G(t) \frac{M_{pi}(t) \times M_{aj}(t)}{R_{ij}(t) + \varepsilon} \left( X_j^d(t) - X_i^d(t) \right), \tag{7}$$

where $R_{ij}(t)$ is the Euclidian distance between two agents i and j, $M_{aj}$ is the active gravitational mass associated with agent j, $M_{pi}$ is the passive gravitational mass related to agent i, $G(t)$ is the gravitational constant at time t, and ε is a tiny constant.

$$R_{ij}(t) = \|X_i(t), X_j(t)\|_2. \tag{8}$$

We assume that the total force acting on agent i in dimension d is a randomly weighted sum of the dth components of the forces exerted by other agents to give our method a stochastic characteristic:

$$F_i^d(t) = \sum_{j=1, j \neq i}^{N} rand_j \, F_{ij}^d(t), \tag{9}$$

where $rand_j$ is a number chosen at random from [0, 1].

As a result, according to the law of motion, the agent's acceleration *i* at time *t* in direction *dth*, denoted by the symbol $a_i^d$, is as follows:

$$a_i^d(t) = \frac{F_i^d(t)}{M_{ii}(t)}, \tag{10}$$

where $M_{ii}$ is the *ith* agent's inertial mass. Additionally, an agent's future velocity is calculated as a portion of its present velocity plus its acceleration. Therefore, the following formulas might be used to determine its location and velocity:

$$v_i^d(t+1) = x\, v_i^d(t) + a_i^d(t), \quad (11)$$

$$x_i^d(t+1) = x_i^d(t) + v_i^d(t+1), \quad (12)$$

where $rand_i$ is a variable with a uniform distribution in the range [0, 1]. This random integer is used to give the search a randomized characteristic. To manage the search accuracy, the gravitational constant, $G$, is first established and then lowered over time. In other words, $G$ is a function of time (*t*) and the beginning value ($G_0$)

$$G(t) = G(G_0, t). \quad (13)$$

The fitness assessment only calculates the masses of gravity and inertia. An agent with a higher bulk is more effective. This indicates that better agents move more slowly and have greater attraction. The values of the masses are determined using the map of fitness under the assumption that the gravitational and inertia masses are equal. These equations allow us to update the gravitational and inertial masses:

$$M_{ai} = M_{pi} = M_{ii} = M_i, i = 1,2,\dots,n, \quad (14)$$

$$m_i(t) = \frac{fit_i(t) - worst(t)}{best(t) - worst(t)}, \quad (15)$$

$$M_i(t) = \frac{m_i(t)}{\sum_{j=1}^{n} m_j(t)}, \quad (16)$$

Where $fit_i(t)$ denotes the agent's *i* fitness value at time *t*, and *worst(t)* and *best(t)* are defined (for a minimization issue) as follows:

$$best(t) = \min_{j \in \{1,\dots,N\}} fit_i(t), \quad (17)$$

$$worst(t) = \max_{j \in \{1,\dots,N\}} fit_i(t). \quad (18)$$

It should be observed that Eqs. (17) and (18) become Eqs. (19) and (20), respectively, for a maximization problem:

$$best(t) = \max_{j \in \{1,\dots,N\}} fit_i(t), \quad (19)$$

$$worst(t) = \min_{j \in \{1,\dots,N\}} fit_i(t). \quad (20)$$

A suitable balance between exploration and exploitation may be achieved, for example, by decreasing the number of agents over time in Eq. (9). Therefore, we suggest that only a group of agents with greater mass exert force on one another. However, we should exercise caution while implementing this approach since it can weaken exploration power and strengthen exploitation capabilities.

It is worth noting that the algorithm must employ exploration from the beginning to prevent getting stuck in a local optimum. Exploration must end after a certain number of repetitions, and exploitation must begin. By limiting exploration and exploitation, the GSA may function better. Only the *Kbest* agents will draw in the other players. *Kbest* is a function of time, starting at *K0* and decreasing over time. In this approach, initially all agents exert force on the others, but as time goes on, *Kbest* progressively decreases, and ultimately just one actor exerts force on the others. As a result, Eq. (9) might be changed to read:

$$F_i^d(t) = \sum_{j \in Kbest, j \neq i} rand_j \, F_{ij}^d(t) \tag{21}$$

where *Kbest* is the group of the first *K* agents with the greatest mass and best fitness value.

### 3.2.1 Variants of GSA

Both the problem variables and the manner in which the agents are represented may be programmed in a range of forms. The five types that are most familiar to the general public are continuous, real value, binary value, discrete, and mixed. Variations of the GSA that have various sorts of illustrations include continuous, real value, binary value, discrete, and mixed variants (Rashedi et al., 2018). Table 5 explains the GSA versions.

**Table 5**
**Variants of GSA**

| Variants | Explanation | Years | Ref. |
|---|---|---|---|
| **Real-Value** | This is the first version of GSA that can be used to address real-world issues. | 2009 | (Rashedi et al., 2009) |
| **Binary** | When an item is at a desirable location near the global optimum, its velocity should be as close to zero as is physically possible. Because of this, Rashedi et al. proposed the binary GSA (BGSA). | 2010 | (Rashedi et al., 2010b) |
| | It was suggested to improve BGSA by changing the movement probability function. | 2014 | (Rashedi & Nezamabadi-Pour, 2014) |
| **Discrete** | An integer vector is used to describe the discrete optimization issue, where new positions are chosen at random from the discrete values available depending on velocity. | 2012 | (Shamsudin et al., 2012) |
| **Mixed** | It was decided to make use of objects that had both binary and continuous variables. As a consequence of this, the equations describing movement change depending on whether the dimension being considered is real or binary. | 2013 | (Sarafrazi & Nezamabadi-Pour, 2013) |
| **Quantum** | Quantum behavior is exhibited by agents in QGSA, which is distinguished by waves of quantum. | 2012 | (Soleimanpour-Moghadam & Nezamabadi-Pour, 2012) |
| **Constraint** | A new version of GSA was created, according to Yadav et al., to address constraint issues. | 2013 | (Yadav & Deep, 2013) |
| **Multimodal** | Yazdani et al. suggests a novel multi-modal optimization framework. The concept of breaking up a core population swarm of masses into smaller sub-swarms and | 2014 | (Yazdani et al., 2014) |

| | maintaining them is covered by NGSA. | | |
|---|---|---|---|
| **Multi-objective** | A MOGSA was developed by Hassanzadeh and colleagues to hold the optimal Pareto solutions in a grid-structured repository. This MOGSA employs an Elitist Policy and a Uniform Mutation Operator. | 2010 | (Hassanzadeh & Rouhani, 2010) |
| **Hierarchy and distributed framework** | Y. Wang et al. provides a GSA with a hierarchy and a distributed framework (DGSA) to improve GSA from the viewpoint of the structure of the population. | 2021 | (Y. Wang, Gao, Yu, et al., 2021) |
| **Multi-layer** | Based on the two-layered structure of the GSA, four levels are created: population, iteration best, personal best, and global best. Dynamically developed hierarchical interactions spanning four levels are used in various search queries to dramatically enhance the population's exploration and exploitation abilities. | 2021 | (Y. Wang, Gao, Zhou, et al., 2021) |

Table 6 gives a link to the source code for the GSA versions.

**Table 6**
**Links to the source code of the Variants of GSA**

| **Variants** | **Link** |
|---|---|
| **Real-Value** | https://github.com/SajadAHMAD1/LCGSA-for-MLP-Training/blob/4621038eabb3c4427931337d157ba643f3c18a8a/Gconstant.m |
| **Binary** | https://github.com/Bahar-nkr/BQPSO/blob/1e567da47d5af26b68afaccba0628b1e2e9784d4/Binitialization.m |
| **Discrete** | https://pan.baidu.com/s/1DL1W1HKDqYsp5sKhXKwiYA |
| **Mixed** | https://github.com/Anjnesh/Research-paper-search-engine/find/6bb402825152c60b8b0b9d444c9ddc554a77e1f3 |
| **Quantum** | Not Available |
| **Constraint** | https://pan.baidu.com/s/1F4BKPkz_6DoWkVfEMVPkqw |
| **Multimodal** | Not Available |
| **Multi-objective** | Not Available |
| **Hierarchy and distributed framework** | https://raw.githubusercontent.com/toyamaailab/toyamaailab.github.io/main/resource/HGSA.zip |
| **Multi-layer** | Not Available |

### 3.2.1.1 Real GSA

In the research paper by (Rashedi et al., 2009), the first version of the real GSA was developed to solve optimization problems involving variables with real values. Up to this point, a large number of improved variations of the real-valued GSA have been described. PSO has been applied to GSA, which is originally memoryless, by multiple researchers (Pei & Haibin, 2012) and (Gu & Pan, 2013) and along with merging (Mirjalili et al., 2012) and (Mallick et al., 2013) and (Precup et al., 2014) and (Jayaprakasam et al., 2015) the memory-based PSO velocity term into the GSA velocity term.

### 3.2.1.2 Binary GSA

There are two possible values for variables in various optimization problems: 0 and 1. A binary version of the heuristic optimization method is utilized to address these issues. This strategy, known as "BGSA," was created by (Rashedi et al., 2010a). In a binary setting, everything has a 0 or 1. The equivalent number goes from 0 to 1, or vice versa, depending on which direction you move in.

### 3.2.1.3 Discrete GSA

The distinctive gravitational search algorithm, or "DGSA," according to (Shamsudin et al., 2012), depicts a discrete optimization problem as an integer vector. Using this method, the acceleration and velocity are calculated similarly to how GSA works.

### 3.2.1.4 Mixed GSA

A hybrid version of GSA that covers items with both binary and continuous properties was introduced by (Sarafrazi & Nezamabadi-Pour, 2013). The movement equations in this version alter depending on whether the dimension is binary or real. It's particularly helpful when addressing situations that include both real and binary variables in their respective formulations (Rashedi et al., 2013).

### 3.2.1.5 Quantum GSA

The creation of the new generation of computers known as quantum theory notions in physics is termed "quantum computing" or "QC". The researchers of sources (Soleimanpour-Moghadam et al., 2014) and (Soleimanpour-Moghadam & Nezamabadi-Pour, 2012) presented a quantum-behaved GSA, or "QGSA". Quantum waves are defined by the behavior of QGSA agents.

### 3.2.1.6 Constraint GSA

As of 2013 (Yadav & Deep, 2013), there is a modern edition of GSA available for dealing with constraint issues. This updated version computes constraint violation via parameter-exempt constraint dealing. In a different research, dynamic constrained optimization difficulties in the reference were addressed using GSA and a modified repair method (Pal et al., 2013).

### 3.2.1.7 Multimodal GSA

Termed Niche GSA "NGSA" is a novel framework for multimodal optimization introduced by (Yazdani et al., 2014). The three methods that are being suggested here are the NGSA, which expands on the idea of conserving smaller sub-swarms of populations by breaking up larger populations, or "swarms," of masses into smaller groups; the K-Nearest Neighbors (K-NN) strategy; the elitism strategy; and a variant of the active gravitational mass idea.

### 3.2.1.8 Multi-objective GSA

For multiobjective optimization "MOGSA," GSA comes in several variations. Rather than a single aim, many real-world situations necessitate the simultaneous optimization of many objectives. Multi-objective problems are what they are called. In situations like this, there exists a group of nondominated options that are collectively referred to as the Pareto optimum solutions. (Hassanzadeh & Rouhani, 2010) suggested using an Elitist Policy and a Uniform Mutation Operator in a grid-structured archive for archiving optimal Pareto results (MOGSA). In another work (Nobahari et al., 2012), an elitism approach was presented and the "NSGSA" nondominated sorting gravitational search algorithm was designed including an additional archive to keep the optimal Pareto solutions. Archived items and moving objects are the two categories of objects.

### 3.2.1.9 GSA with Hierarchy and Distributed Framework (DGSA)

(Y. Wang, Gao, Yu, et al., 2021) provide a GSA with a hierarchy and a distributed framework (DGSA) to improve GSA from the standpoint of population dynamics. A distributed framework creates hierarchical subpopulations within a population. Any hierarchical subpopulation has three tiers: the bottom, middle, and top. Subpopulation communication improves population variety and the caliber of solutions.

### 3.2.1.10 Multi-layer

Despite being a strong population-based algorithm, GSA performs poorly in searches and converges too quickly. (Y. Wang, Gao, Zhou, et al., 2021) suggests a multilayered GSA known as the MLGSA to address these problems. Inspired by the two-layered structure of GSA, four levels are created: population, iteration-best, personal-best, and global-best. To greatly improve the population's exploration and exploitation abilities, hierarchical interactions were dynamically created across four levels in a variety of search phrases.

### 3.2.2 Various operators of GSA

In certain cases, innovative operators were created or current operators were reworked to provide GSA with more capabilities for exploitation and exploration. The new GSA operators are Disruption, Mutation, Chaotic, Crossover, Escape, Black Hole, and Kepler. The operators of GSA are explained in Table 7.

**Table 7**
**Operators of GSA**

| Operators | Explanation | Years | References |
|---|---|---|---|
| **Disruption** | A novel astrophysically inspired operator with the snappy name "Disruption" is presented to expand the core Gravitational Search Algorithm's capabilities in terms of exploration and exploitation. | 2011 | (Sarafrazi et al., 2011) |
| **Mutation** | The Non-dominated Sorting GSA (NSGSA) is a novel multi-objective form of GSA suggested by the authors. In this regard, two novel mutation operators, named reordering mutations and signs, have also been suggested. | 2012 | (Nobahari et al., 2012) |
| **Chaotic** | To expand the flexibility of GSA, more operators should be introduced. To address premature convergence, researchers presented a modified gravitational search algorithm (MGSA) that introduces an innovative chaotic operator. | 2012 | (Han & Chang, 2012) |
| **Crossover** | To handle more challenging issues, the Neighborhood Crossover Operator (NCO), also known as the Gravitational Search Algorithm (GSA), was employed with the GSA to increase its exploration and use. | 2013 | (Shang, 2013) |
| **Escape** | The recommended technique combines the benefits that are associated with two different methods of image segmentation, namely grouping, and region-growing. The gravitational image segmentation algorithm (SGISA) has recently included a new operator that goes by the name "escape". The operators is based on the physical idea of escape velocity. | 2013 | (Rashedi & Nezamabadi-Pour, 2013) |
| **Blackhole** | Whenever a star is turned into a black hole, it has incredibly high gravity that prevents anything from escaping, and things that are close to the black hole are subjected to a powerful force known as tidal force, which drives them to collapse into the black hole. Researchers proposed a novel operator (BH-GSA) that incorporates these properties and is hybridized with GSA to avoid premature convergence and increase GSA's exploration and exploitation capabilities. | 2014 | (Doraghinejad & Nezamabadi-pour, 2014) |
| **Kepler** | The GSA operator was replaced with the Kepler operator, which is a high-level relay hybrid algorithm, because both the GSA and Kepler algorithms are self-contained as well as performed inside the prescribed order. | 2015 | (Sarafrazi et al., 2015) |

## 5. Applications of KH and GSA in Healthcare

The prevention, diagnosis, treatment, recovery, or cure of disease, sickness, injuries, and other physical and mental disorders in humans are all components of what is collectively referred to as healthcare. Professionals in the medical area and those working in fields closely related to healthcare are the ones that provide healthcare. In the time of technology, it has been able to benefit from algorithms in most of the fields of our daily lives. Moreover, directly or in hybrid with other algorithms, they have become a means of facilitating the management of many important secrets that are related to human life. The krill herd algorithm and gravitational search algorithm are two successful algorithms for tackling real-world issues in various medical disciplines, including healthcare. For the information to be clear, this part has been divided into two subsections:

### 4.1 Application of KH in Healthcare

One of the sectors that have benefited the most from recent advances in computational technology, including software and hardware, is medicine. The krill herd algorithm has received a lot of interest and has been used in a variety of applications.

Sentiment analysis is required for the processing of subjective data as well as the capacity to discriminate between feelings and emotions expressed on social media. The study of people's views, sentiments, and attitudes toward various topics of interest is known as sentiment analysis or opinion mining. Processing of natural languages, classification of text, retrieval of information, computational linguistics, and data mining are all used in this process. This also entails activities like sentiment analysis, extraction, and categorization on a variety of resources, including the Internet, discussion forums, blogs, websites, and social networks, among others. As a result of this, sentiment analysis is employed by a wide range of industries, including marketing, politics, tourism, and healthcare. However, despite its numerous advantages, the healthcare domain has received little attention. Using sentiment analysis, it is possible to learn how people think and feel about medical practitioners, medications, ailments, and treatments. This data may be used to improve patient treatment, public health monitoring, and epidemic control, to mention a few examples (Ramírez-Tinoco et al., 2019) and (Htet et al., 2019).

From another perspective, on sevral medical websites and forums, individuals discuss their health problems and trade details about their ailments, symptoms, and prescription drugs. (L. M. Q. Abualigah, 2019) presented a model using sentiment analysis for text document clustering (TDC) using the KH algorithm. They gained from employing the Krill Herd algorithm for grouping text data in sentiment classification throughout healthcare for data extraction and analysis (Rafi & Bharathi, 2019). Furthermore, (Preethi, 2018) using Recommender Systems "RS" is critical in providing users with appropriate suggestions. To acquire improved and ideal results, the optimization approach is essential for optimizing the variables of the selected algorithm. The method was developed using an improved krill herd algorithm to optimize and increase the accuracy of recommendations. The upgraded k-means clustering technique was used to group diabetes patients' profiles, and the improved krill herd optimization process was used to fine-tune the findings.

Coupled with this, the theory of KH is efficient and capable in the fields of classification and selecting features from medical images. (Rafi & Bharathi, 2019) Breast cancer medical data mining and classification uses a hybrid Adboost KNN with the krill herd algorithm. (Abugabah et al., 2021) the Krill Herd algorithm is used to capture and evaluate many aspects of the brain signal to choose the best characteristics. (Kumari & Arumugam, 2015) A Krill herd optimization approach was utilized to provide a more thorough categorization of the dataset of breast cancer. It is also proposed to develop the best classification criteria by the use of a hybrid krill herd algorithm termed "HKH". According to (Kalyani et al., 2021), the krill herd algorithm used in the segmentation of medical images "CSA," the effectiveness of the suggested work is examined on three different medical images at four, five, six, and seven threshold values, and it is compared to contemporary algorithms like Krill herd, the Cuckoo search algorithm, and Teaching-learning based optimization (TLBO). The author describes an approach for extracting fuzzy rules from the Breast Cancer dataset (Mohammadi et al., 2014). To extract fuzzy rules, the Krill Herd Algorithm, an imitation-based evolutionary method, is utilized.

Moreover, to overcome the limitations of conventional color picture segmentation algorithms, (He & Huang, 2020) presented a powerful krill herd technique to identify the appropriate threshold values derived from empirical metrics, such as Otsu, Kapur, and Tsallis entropies. Three bio-inspired optimization methods have been implemented in images of medical devices, and the median error of resolution by iteration was compared with the Shannon entropy as an assessment function (Wachs Lopes et al., 2018). To evaluate the results of the three multilevel segmentation algorithms with one, two, and three thresholds, the assessments were performed on two distinct medical image datasets. Additionally, in response, (Devi et al., 2020) showed how to develop and use a hybrid krill herd extreme learning machine "ELM" neural network classifier. This classifier is capable of identifying two-dimensional MRI brain cancer pictures as belonging to one of the following four categories: normal, astrocytoma, meningioma, or oligodendroglia.

In a way, medical imaging is critical in determining a patient's health status and providing appropriate therapy. Although medical images are increasingly being utilized to treat and assess a broad range of disorders, their intrinsic complexity makes them challenging to diagnose because of the various overlapping components. Medical images including X-rays, MRIs, mammograms, and computed tomography (CT) scans lack sufficient details to provide an accurate diagnosis because of a variety of issues, including inadequate lighting conditions, background noise, imaging equipment technology limits, and other factors. The authors describe a new krill herd-based architecture for medical image contrast and sharpness improvement that takes use (Kandhway et al., 2020).

**4.2 Application of GSA in Healthcare**

(Jayashree & Ananda Kumar, 2019) the importance of diabetes was identified at an early stage by building an expert system because of its importance. All through this method, the expert system faces several difficulties, including low prediction accuracy because of the diabetic feature's large size, which affects the system's overall efficiency. The evolutionary correlated gravitational search algorithm (commonly known as "ECGS") was developed as a direct result to choose the most desired traits. Any diabetes feature is assessed based on the association and mutual information, with the lowest amount of computer costs and time adopted. For effective diabetes-related trait prediction, a genetically adapted Hopfield neural network, or "GHNN," analyzes a set of predetermined features. The main goal of this study is to create an expert system that can diagnose diabetes. The functionality of the proposed framework is divided into two components. In the first stage, the "ECGS" is utilized to collect diabetes data from patients and also to minimize the feature set's dimensionality. The whole diabetes identification procedure is improved by this algorithm's effective feature extraction method. In the second step of the process, diabetes is categorized using a genetically optimized Hopfield neural network, or "GHNN". In the augmented genetic algorithm, GHNN is employed as a classifier. This genetic algorithm is crucial in the fight against disease detection errors by decreasing the computing time and cost while significantly increasing diabetes category accuracy. This comparative study utilizes a variety of performance evaluation criteria, including sensitivity, accuracy, specificity, and error rate. The suggested expert system's experimental findings are compared with other current systems to assess its performance and accuracy. (Beevi et al., 2017) recommended a localized dynamic form show based on the Krill Herd Algorithm to isolate cell cores from the foundation, and after that, utilized a multi-classier framework based on a profound belief that we organize to categorize cells into mitotic and non-mitotic bunches.

(González-Álvarez et al., 2013) proposed two distinct swarm intelligence algorithms, Artificial Bee Colony "ABC" and Gravitational Search Algorithm, to solve the Motif Discovery Problem ("MDP"), a DNA sequence analysis difficulty. The MDP is a specific kind of NP-hard optimization problem. Furthermore, the primer design task is considered an optimization problem. (Amoozegar & Rezvannejad, 2014) outlined a plan for enhancing the outcomes of previous attempts. A more efficient solution is necessary because the problem of primer design has such a large and convoluted number of nodes. A heuristic algorithm that imitates how natural systems carry out their tasks is the gravitational search algorithm. Their success in handling search and optimization issues has garnered a significant amount of attention due to their achievements in this area.

Clustering plays a significant part in many data-mining operations, such as information retrieval, text mining, and image segmentation. A crucial component of medical image processing is clustering. As a consequence of the vast amount of data, manual segmentation is time-consuming and complicated, and the results are susceptible to errors. As a result, automated segmentation systems are becoming increasingly important in today's world. In this study, an automated method for segmenting cerebrospinal fluid, white matter, and grey matter from MRI images of the human brain is described. Similarly, (Hooda & Verma, 2022) presented a new clustering method for MRI brain picture segmentation called the Fuzzy-Gravitational Search Algorithm.

Additionally, the GSA, a very potent meta-heuristic algorithm, exhibits new levels of classification accuracy. The use of a gravitational search algorithm to analyze medical data for disease prediction necessitates the use of effective feature selection strategies(Nagpal et al., 2017). In a separate study (Dash et al., 2015), they proposed a hybridization model of two metaheuristics: gravitational search algorithm and particle swarm optimization. This process of segmenting the medical data was carried out using the outcome model. To detect tumors in the kind of breast cancer in mammography pictures, (Shirazi & Rashedi, 2016) suggested a technique by combining a Support Vector Machine (SVM) and Gravitational Search Algorithm (MGSA). The newly suggested technique was successful in cutting down on the total number of features while simultaneously increasing the SVM classifications' degree of precision. In another work, (Lakshmanaprabu et al., 2019) evaluated CT scans of lung images using the Optimal Deep Neural Network (ODNN) and Linear Discriminate Analysis (LDA). To differentiate between malignant and benign lung nodules, LDA is used to reduce the dimensionality of deep features extracted from CT lung images. The ODNN's capacity to identify lung cancer from CT scans is improved by the Modified Gravitational Search Algorithm (MGSA). Meanwhile, a training technique for updating network parameters based on a combination of the gravitational search algorithm and the error backpropagation algorithm (EBPA) (Chao et al., 2018) is suggested. Likewise, medical image classification is important in both clinical and research settings.

(Rajesh Sharma & Marikkannu, 2015) The hybrid approach supports vector machines and a refined gravitational search algorithm. For the categorization of brain cancers using magnetic resonance imaging, the best characteristics are identified. The suggested RGSA is used to choose the best features. The goodness of the classifier technique is evaluated using SVM, backpropagation networks, and k-nearest neighbor. A unique hybrid wrapper technique that combines the features of a teaching-learning-based algorithm with a gravitational search algorithm is described in another work under the name TLBOGSA (Shukla et al., 2020). In addition, TLBOGSA incorporates a novel encoded approach, which enables the transformation of the incessant search area into the binary search space and the formation of binary TLBOGSA. Its suggested technique begins with a feature selection based on the principle of minimal redundancy maximum relevance (mRMR), to select essential genes from gene expression datasets. The genes that contain significant information are then selected using the wrapper technique on the mRMR-reduced data. To improve the search's capabilities via the process of evolution, they employed a gravitational search algorithm throughout the teaching phase. The proposed method makes use of a Naïve Bayes classifier as a fitness function to choose the genes that are the most prudent in terms of their potential contribution to the accurate categorization of cancer. As a result, (Al-Ma'aitah & AlZubi, 2018) propose the Gravitational Search Optimized Echo State Neural Networks for effectively predicting oral cancer. X-ray pictures from the oral cancer database were used first. After that, the region that had been influenced was segmented using an upgraded version of the Markov Stimulated Annealing algorithm, and the characteristics were retrieved from the region that had been segmented. The suggested classifier is used to assess the generated characteristics. (Kate & Shukla, 2022) used GSA to describe the automated categorization of mammographic breast tissue density, which is very vital in morphological analysis for anomaly diagnosis. Pre-processing, mammography enhancement, and categorization are all part of their proposed research work. The foreground picture is extracted from the source image using image pre-processing. The extraction of images is a well-known optimization challenge.

In recent times, COVID-19 disease has become a new widespread disease in the whole world. In the diagnosis process, detecting desire is difficult and inaccurate. (Ezzat et al., 2021) proposed a method named GSA-DenseNet121-COVID-19, constructed with a hybrid gravitational search algorithm and convolutional neural network (CNN). The GSA is put to use to locate the hyperparameters of the DenseNet121 design that have the optimum values. To aid in the identification of COVID-19 utilizing chest X-ray images, this architecture has to have a high degree of precision. Moreover, a more efficient healthcare system is also the main problem that the world's growing population poses in the modern era. An effort called the Internet of Medical Things (IoMT) aims to create a more thorough and ubiquitous health monitoring system. (Sampathkumar et al., 2022) describes how to develop a metaheuristic optimization approach for massive data processing in the IoMT by combining the gravitational search optimization algorithm (GSOA) and a reflective belief network with convolutional neural networks (DBN-CNNs). The proposed algorithm has been classified to forecast the normal and abnormal ranges of diabetes. Based on the classified results, SVM was used to estimate the risk of a heart attack. Because of this type of demand, healthcare supply chains must maintain high inventory levels to ensure that medications are always available to save people's lives. They devised a system for efficient inventory control that makes use of data mining ideas as well as the gravitational search algorithm (Arul Valan & Baburaj, 2020). Employing epidemiology data, (Choudhry et al., 2022) created Enhanced Gravitational Search Optimization with a Hybrid Deep Learning Model (EGSO-HDLM) for COVID-19 diagnosis. The primary goal of developing the EGSO-HDLM model was to identify and classify COVID-19 through epidemiological data. The modified GSA was employed in monitoring the concentration of plasma glucose, which is critical for early diagnosis of diabetes and treatment of diabetes. Early illness detection provides for improved disease control (Kartono et al., 2022).

## 6. Modifications of KH and GSA Used in Healthcare

An altered version of the Krill Herd Algorithm has been created to improve the precision of the suggested goods (Devi et al., 2020). MMKH is a text clustering approach that also includes a hybrid function.

## 7. Hybridization of KH and GSA Used in Healthcare

Hybridization of the two algorithms is a typical strategy for gaining the advantages of both while minimizing their drawbacks. The most typical hybridization combines a classifier with Nature Inspired

Algorithms. The most common tasks for which Nature Inspired Algorithms Feature Selection (NIA-FS) are adapted and used are microarray and medical applications (Khurma et al., 2022). The hybridization of KH and GSA, on the other hand, may be separated into the following subparts. The hybridization of KH and GSA is explained in Table 3.

### 6.1 Hybridization of KH

(Rafi & Bharathi, 2019) used the KH to determine the optimal features. A hybrid Adaboost-KNN classification algorithm was used to classify the optimal features. The results obtained from the hybrid algorithms achieve the best accuracy, sensitivity, and specificity. Both global and local characteristics are extracted from brain cancer images using Krill Herd and Extreme Learning Machines (Preethi, 2018). This hybrid method enhances the classification results while also cutting down on computing time and increasing accuracy. To improve cluster search choice, a hybrid Kill Herd algorithm with the Tabu search is given (P.T.Karthick, 2019). When a global decrease of total Gibbs energy is necessary, the KH and the modified Lévy-Flight Krill Herd algorithm (LKH) have been employed to determine phase stability (PS) and phase equilibrium (PE) in non-reactive (PE) and reactive (rPE) systems (Moodley et al., 2015). Additionally, the topic of combined heat and power economic dispatch (CHPED) was approached using the krill herd technique (Adhvaryyu et al., 2014). Also, Hybrid heuristic algorithms are being developed to solve the main shortcomings of traditional optimization algorithms, such as getting stuck in local optima and experiencing performance deterioration in real-world scenarios with complicated issues(Damaševičius & Maskeliūnas, 2021). In order to prevent all agents from being stuck in subpar local optimum zones, the application of the suggested strategy makes an effort to consider the benefits of the (KH) and the (ABC)(Damaševičius & Woźniak, 2017). In Table 8, the hybridization of the theory was discussed in detail.

**Table 8**
**Hybridization of KH**

| Author | Proposed Algorithm | Description | Application | Results | Ref. |
| --- | --- | --- | --- | --- | --- |
| Dudekula Mahammad Rafi and Chettiar Ramachandra Bharathi | Hybrid Adaboost KNN | It is to determine the optimal feature | Classification | Algorithms produced the best results in terms of accuracy, specificity, and sensitivity. | (Rafi & Bharathi, 2019) |
| J. Preethi | Krill Herd-Extreme Learning Machine (ELM) Network | For extraction of features from brain cancer images. | Classification and Extraction | The proposed hybrid algorithm improves classification results while reducing computation time and improving accuracy. | (Preethi, 2018) |
| Ahamad Tajudin Khader, Laith Mohammad Abualigah, Amir Hossein Gandomi, Mohammed Azmi Al-Betar | H-KHA is a combination of the KH algorithms and the harmony search (HS) method. | It is used in the data clustering field. | Classification | It is proactive and effective in addressing clustering issues. | (L. M. Abualigah et al., 2017) |

## 6.2 Hybridization of GSA

The process of hybridizing algorithms may result in improvements to the overall algorithm's exploration and exploitation (Sheikhpour et al., 2013). The algorithm's lack of accuracy may be improved, in particular, by hybridization with a search engine strategy that focuses on the outcomes. PSO and GSA algorithms were combined for the categorization of medical datasets by (Dash et al., 2015). Five medical datasets from the UCI source were classified using the GS and PSO based on Fuzzy MLP, and the outcomes were then contrasted with those of the suggested technique. (Dash et al., 2015). The GS and PSO based on Fuzzy MLP were utilized to classify five medical datasets from the UCI source, and the results were then compared to those of the proposed method. From another aspect, (Shirazi & Rashedi, 2016) proposed the algorithm SVM and mixed GSA (MGSA) for application to detect and classify breast cancer tumors. The GSA-DenseNet121-COVID-19 (Ezzat et al., 2021) combines CNN and GSA to detect COVID-19 illness from chest x-ray images (Ezzat et al., 2021). In another work (Sampathkumar et al., 2022), they introduced the gravitational search optimization algorithm and a reflective belief network combined with convolutional neural networks for classifying normal and abnormal diabetes. Meanwhile, (Chao et al., 2018) presented a technique that trained the network to modify its variables using a mix of the error backpropagation approach (EBPA) and the gravitational search algorithm. (Rajesh Sharma & Marikkannu, 2015) developed a hybrid strategy that included a refined gravitational search algorithm with support vector machines to extract characteristics from medical images in MRI and classify brain tumors. Also, (Shukla et al., 2020) presented a hybrid wrapper method that combines aspects of teaching learning-based algorithms with gravitational search algorithms for feature selection and gene discovery for various forms of cancer. In the sequential flow shop setting, the challenge of scheduling *n* independent work with varying due times on two machines is explored. To address this issue, (Pirozmand et al., 2022) present SA-GSA, a hybrid Simulated Annealing with the Gravitational Search Algorithm. Given the ease with which GSA can fall into local optimum traps and the necessity to increase search accuracy, a fusion method combining gravitational search and tabu search (TS-GSA) has been developed (Lou, 2019). Although it can efficiently optimize many problems, it is prone to early convergence and has a limited search capability. To address these limits, a gravitational research approach with a hierarchy and a distributed architecture was devised. A distributed framework gathers various subpopulations at random and maintains them using a three-tiered hierarchy. Finally, communication among subpopulations improves search performance (Y. Wang, Gao, Yu, et al., 2021). Table 9 gives accurate details about the hybridization of GSA.

**Table 9**
**Hybridization of GSA**

| Author | Proposed Algorithm | Description | Application | Results | Ref. |
|---|---|---|---|---|---|
| Tirtharaj Dash et al. | Fuzzy MLP-GSAPSO | For the classification of the medical database. | Classification | The categorization accuracy is improved with the help of the suggested model. | (Dash et al., 2015) |
| Fatemeh Shirazi and Esmat Rashedi | Support vector machine hybrid with gravitational | The purpose of this is to classify breast | Classification | The suggested approach enhances classification accuracy while reducing the number of | (Shirazi & Rashedi, 2016) |
| | search algorithms | cancer tumors. | | characteristics. | |
| Dalia Ezzat et al. | Hybrid CNN and GSA | It has yet to be determined COVID-19 disease. | Classification | The proposed model improves the accuracy of a classification and is capable of diagnosing COVID-19. | (Ezzat et al., 2021) |
| Sampathkumar et al., | Combine GSOA and DBN-CNNs | For classification and prediction of diabetes disease. | Classification and Prediction | The proposed methods for diabetes disease classification and prediction. | (Sampathkumar et al., 2022) |
| Zhen Chaoa, Dohyeon Kima, and Hee-Joung Kima | Hybrid GSA and EBPA | For the classification of CT scan images. | Classification | To achieve a higher degree of precision in the training of the neural network. | (Chao et al., 2018) |
| R. Rajesh Sharma and P.Marikkannu(Rajesh Sharma & Marikkannu, 2015) | Hybrid RGSA and SVM | It is for determining optimal features and classification. | Classification and Extraction | The proposed algorithm provides how to extract and choose features systematically and efficiently. | (Rajesh Sharma & Marikkannu, 2015) |
| Alok Kumar Shukla et al. | Combine TLBO and GSA | It is for determining optimal features and classification. | Classification and Extraction | It is for classification accuracy and the best amount of feature sets. | (Shukla et al., 2020) |
| Abdolreza Hatamlou, SalwaniAbdullah, HosseinNezamabadi-pour | Combine GSA and K-means(GSA-KM) | It is employed in the clustering of data items. | Clustering | The GSA-KM algorithm accelerates the convergence of the GSA technique while assisting the k-means algorithm in escaping from local optima. | (Hatamlou et al., 2012) |

## 8. Limitation of Algorithms

All algorithms have good sides but are not equally empty of drawbacks and weaknesses. In this section, the main limitations of KH and GSA are explained.

## 7.1 Limitation of KH

Despite KH's great performance, local optima stalling and sluggish convergence speed are two common issues while handling difficult optimization tasks. Moreover, the KH algorithm shines at local exploitation but challenges global exploration, particularly when dealing with high-dimensional multimodal function optimization, because the approach does not always converge quickly. For a more solid implementation and more dependable results, additional study on the theoretical analysis of KH is required. New optimization techniques may also be added to the fundamental KH.

## 7.2 Limitation of GSA

GSA has the advantages of being simple to construct, converging quickly, and having a minimal computing cost. Nevertheless, GSA has the disadvantage that it is easy to choose a local optimal solution and that as the search goes on, its convergence speed slows down. This is an algorithmic constraint. However, the search capabilities of GSA are restricted when dealing with complicated issues that include several local optimum solutions. To put it another way, computing GSA requires a significant amount of time. Aside from that, the main problem with GSA, just like with other heuristic algorithms, is that it tends to converge too quickly. Additionally, it the difficulty of choosing the appropriate gravitational constant parameter G using complex operators. Moreover, large-scale optimization and dynamic challenges in particular remain open for discussion.

## 9. Conclusions and Future Work

In conclusion, the study concludes by summarizing the most recent developments in AI applications, particularly in the healthcare and medical fields, during the last seven years, with a focus on the gravitational search algorithm and the krill herd algorithm. Many optimization issues are solved with the new KH and GSA algorithms.

Above all, the findings show that the KH and GSA algorithms are quite effective at tackling a variety of issues. Furthermore, it outperforms several current methods. There may be fewer parameters to fine-tune in KH than in other algorithms. From another perspective, the GSA works very carefully in the field of classification and segmentation of medical images. Furthermore, the time computational complexity of KH is $O(n^4)$ greater than that of GSA, which is related to $O(n^2)$. As a consequence of this, it became clear that the krill herd algorithm and the gravitational search algorithm are those algorithms that are now being used in the fields of science and engineering to find solutions to issues that occur in the real world. The field of medicine is significantly benefited from the contribution of these algorithms, which play an active role.

Moreover, From the results of the study, we can see that metaheuristic algorithms can play a role in solving all real-world problems. They play a significant role, especially in medicine. In the medical field, they try in the following ways: sentiment analysis, document clustering (TDC), breast cancer medical data mining and classification, classification, feature selection (FS), DNA sequence analysis, predicting oral cancer, diagnosis of COVID-19 disease, Internet of Medical Things (IoMT), and classification to forecast the normal and abnormal ranges of diabetes.

The limitations of the survey: the first limitation is that it only selects papers from the years 2015 to 2023. The second is that the articles were published in the most impactful journals and up-to-date conferences. The purpose of this restriction was only to use recent research in the field as the main and reliable sources. Also, avoid selecting many papers published in local journals or fake journals.

In future work, When tackling novel issues, choosing the KH parameters remains a difficult task. Another concern is that the KH algorithm was primarily researched for single-objective situations. In light of this, further research on multi-objective and many-objective optimization using KH may be beneficial.

Despite these attempts, GSA still has a lot of unresolved issues. By defining additional operators or releasing enhanced versions, GSA families may expand. Up to this point, GSA-specific operators included Kepler, Escape, Disruption, and Black Hole. A lot of novel operators might be created, particularly if they draw inspiration from physics ideas on gravity, antigravity, relativity, star clusters, and planetary motions. researchers should work on hybridizing these two algorithms and using them in the field of medicine. This suggestion is to improve the ability and efficiency of these algorithms. Despite this, in many medical fields, it is still manual work and traditional techniques that can cause a lot of inaccuracies and more computational time and effort.

**Conflict of interest:** The authors declare no conflict of interest to any party.

**Ethical Approval:** The manuscript is conducted in the ethical manner advised by the targeted journal.

**Consent to Participate:** Not applicable

**Consent to Publish:** The research is scientifically consented to be published.

**Funding:** The research did not receive specific funding but we want to contribute to a traditional publishing model.

**Competing Interests:** The authors declare no conflict of interest.

**Availability of data and materials:** Data can be shared upon request from the corresponding author.

**Acknowledgment:** None.

along with krill herd algorithm (KHA). *Ingenierie Des Systemes d'Information*, *24*(1), 77–81. https://doi.org/10.18280/isi.240111

Rajesh Sharma, R., & Marikkannu, P. (2015). Hybrid RGSA and Support Vector Machine Framework for Three-Dimensional Magnetic Resonance Brain Tumor Classification. *Scientific World Journal*, *2015*. https://doi.org/10.1155/2015/184350

Ramírez-Tinoco, F. J., Alor-Hernández, G., Sánchez-Cervantes, J. L., Salas-Zárate, M. del P., & Valencia-García, R. (2019). Use of sentiment analysis techniques in healthcare domain. *Studies in Computational Intelligence*, *815*, 189–212. https://doi.org/10.1007/978-3-030-06149-4_8

Rashedi, E., & Nezamabadi-Pour, H. (2013). A stochastic gravitational approach to feature based color image segmentation. *Engineering Applications of Artificial Intelligence*, *26*(4), 1322–1332. https://doi.org/10.1016/J.ENGAPPAI.2012.10.002

Rashedi, E., & Nezamabadi-Pour, H. (2014). Feature subset selection using improved binary gravitational search algorithm. *Journal of Intelligent & Fuzzy Systems*, *26*(3), 1211–1221. https://doi.org/10.3233/IFS-130807

Rashedi, E., Nezamabadi-pour, H., & Saryazdi, S. (2009). GSA: A Gravitational Search Algorithm. *Information Sciences*, *179*(13), 2232–2248. https://doi.org/10.1016/j.ins.2009.03.004

Rashedi, E., Nezamabadi-Pour, H., & Saryazdi, S. (2010a). BGSA: Binary gravitational search algorithm. *Natural Computing*, *9*(3), 727–745. https://doi.org/10.1007/s11047-009-9175-3

Rashedi, E., Nezamabadi-Pour, H., & Saryazdi, S. (2010b). BGSA: Binary gravitational search algorithm. *Natural Computing*, *9*(3), 727–745. https://doi.org/10.1007/S11047-009-9175-3/TABLES/16

Rashedi, E., Nezamabadi-Pour, H., & Saryazdi, S. (2013). A simultaneous feature adaptation and feature selection method for content-based image retrieval systems. *Knowledge-Based Systems*, *39*, 85–94. https://doi.org/10.1016/J.KNOSYS.2012.10.011

Rashedi, E., Rashedi, E., & Nezamabadi-pour, H. (2018). A comprehensive survey on gravitational search algorithm. *Swarm and Evolutionary Computation*, *41*(January), 141–158. https://doi.org/10.1016/j.swevo.2018.02.018

Rodrigues, D., Pereira, L. A. M., Papa, J. P., & Weber, S. A. T. (2014). A binary krill herd approach for feature selection. *Proceedings - International Conference on Pattern Recognition*, 1407–1412. https://doi.org/10.1109/ICPR.2014.251

Sabri, N. M., Puteh, M., & Mahmood, M. R. (2013). A review of gravitational search algorithm. *International Journal of Advances in Soft Computing and Its Applications*, *5*(3).

Sampathkumar, A., Tesfayohani, M., Shandilya, S. K., Goyal, S. B., Shaukat Jamal, S., Shukla, P. K., Bedi, P., & Albeedan, M. (2022). Internet of Medical Things (IoMT) and Reflective Belief Design-Based Big Data Analytics with Convolution Neural Network-Metaheuristic Optimization Procedure (CNN-MOP). *Computational Intelligence and Neuroscience*, *2022*, 1–14. https://doi.org/10.1155/2022/2898061

Sarafrazi, S., & Nezamabadi-Pour, H. (2013). Facing the classification of binary problems with a GSA-SVM hybrid system. *Mathematical and Computer Modelling*, *57*(1–2), 270–278. https://doi.org/10.1016/J.MCM.2011.06.048

Sarafrazi, S., Nezamabadi-Pour, H., & Saryazdi, S. (2011). Disruption: A new operator in gravitational search algorithm. *Scientia Iranica*, *18*(3), 539–548. https://doi.org/10.1016/J.SCIENT.2011.04.003